\documentclass[journal]{IEEEtran}
\usepackage{tabularx}
\usepackage{booktabs}

\usepackage{graphicx}
\usepackage{authblk}

\usepackage{cite}

\ifCLASSINFOpdf
\else
\fi
\usepackage{amsmath}
\usepackage{algorithmic}

\usepackage{stfloats}

\usepackage{xcolor}
\usepackage[normalem]{ulem}

\usepackage{threeparttable}
\usepackage{placeins}

\begin{document}
%
\title{Multi-objective multi-generation Gaussian process optimizer for 
design optimization}
%
%
%



\author[1]{Xiaobiao Huang\thanks{xiahuang@slac.stanford.edu}}
\author[1,2]{Minghao Song}
\author[1]{Zhe Zhang}
\affil[1]{SLAC National Accelerator Laboratory, Menlo Park, CA 94025}
\affil[2]{Illinois Institute of Technology, Chicago, IL 60616.}

\date{\today}
%
%


\markboth{SLAC-PUB-17451, \today}
{}
%



\maketitle

\begin{abstract}
We present a multi-objective evolutionary optimization algorithm that uses 
Gaussian process (GP) regression-based models to select trial solutions 
in a multi-generation iterative procedure. 
In each generation, a surrogate model is constructed for each objective function 
with the sample data. 
The models are used to evaluate solutions 
and to select the ones with a high potential before they are 
evaluated on the actual system. 
Since the  trial solutions selected by the GP models tend to have better performance 
than other methods that only rely on random operations, the 
new algorithm has much higher efficiency in exploring the parameter space. 
{Simulations with multiple test cases 
show that the new algorithm has a substantially higher convergence speed and stability than  
NSGA-II, MOPSO, and some other more recent algorithms.} 
\end{abstract}


%
\IEEEpeerreviewmaketitle

%
%
%
%

\section{\label{sec:intr}Introduction}
\IEEEPARstart{T}{he} 
design of a complex system often requires the search of the ideal solution 
among a multi-variable parameter space. 
The ideal solution may involve a trade-off of competing 
performance requirements. 
In recent years, multi-objective evolutionary algorithms (MOEAs) have been widely adopted to 
discover the set of solutions with the best performances, i.e., 
the Pareto front. These include multi-objective 
genetic algorithms (MOGA)~\cite{DebGAbook,NSGA2,MOEAD} and 
multi-objective particle swarm optimization (MOPSO)~\cite{Kennedy488968,PSO985692Kennedy,MMOPSO,WOFSMPSO2018}.

In the particle accelerator field, there are many challenging design optimization problems, such as lattice design for synchrotron light sources~\cite{YangEPAC08MOGA,BorlandPAC09DA,HUANG201448PSO}, beamline design for photoinjectors~\cite{BazarovMOGA}, and cavity design for superconducting Radio Frequency (SRF) components. 
Numeric optimization is often used in the design studies. 
In a design optimization study, many trial solutions will be evaluated, and typically an evaluation 
involves the numeric simulation of the physics processes that affect the 
system performance. 
Such a simulation could be computationally expensive (e.g., hours), especially as the 
current trend is to build in as many details into the physics model as possible. 
On the other hand, accelerator projects often have tight schedules, with limited time available for design studies. 
Therefore, high efficiency of the optimization algorithm is crucial.
The fast convergence requirement means that the optimizer not only needs to be able to converge to the true Pareto front with a small 
number of evaluations, but also has a small computation overhead. 
In addition, the performance of the optimizer needs be stable in order to avoid the need to rerun the same 
optimization multiple times. 

Both MOGA~\cite{BazarovMOGA,YangEPAC08MOGA,BorlandPAC09DA} and 
MOPSO~\cite{PANG2014124PSO,HUANG201448PSO} algorithms have found use in  accelerator design studies 
in recent years. In MOGA and MOPSO algorithms, an iterative process is executed to 
update a population of solutions. During each iteration, which may be referred to 
as a generation, new trial solutions are generated and evaluated. 
Both methods employ stochastic operations to produce new solutions with existing 
good solutions, although the details differ. 
These operations are heuristically effective, but are intrinsically inefficient as 
the new solutions are not based on any valid prediction. There is a strong incentive to develop more powerful 
methods as the design of future accelerators is becoming more challenging.


Several novel techniques are adopted in some of the more recently developed MOEAs, such as objective decomposition (MOEA/D~\cite{MOEAD}), multiple search strategies (MMOPSO~\cite{MMOPSO}) and problem transformation scheme (WOF-SMPSO~\cite{WOFSMPSO2018}), to tackle complex or large-scale multi-objective optimization problems (MOPs).
These new algorithms can be considerably faster than the conventional MOGA or MOPSO algorithms. However, like the MOGA and MOPSO, the new algorithms do not make full use of the information 
in the evaluated solutions to assist the search for the Pareto front. 


Surrogate assisted MOEAs~\cite{MOEADEGO,KRVEA,ParEGO} have been developed to improve 
the efficiency of the algorithms. 
These algorithms are often based on posterior Gaussian process 
(GP)~\cite{Kushner_1964GP,ZhilinskasGP,Mockus77GP,Jones1998GP, RasmussenGP,  BrochuGPTut2010} models. A posterior Gaussian process is a non-parametric, 
analytic 
model derived from a prior Gaussian process and the sample data, based on  
the Bayesian inference. It serves as a surrogate model of the 
actual physics model that represents the system and is used to produce the sample data. 
The GP model can be used to predict the performance of solutions not yet 
evaluated, along with an uncertainty estimate. 

MOEA/D-EGO~\cite{MOEADEGO} and ParEGO~\cite{ParEGO} are two of the GP assisted algorithms that have excellent performance for many optimization problems. 
However, the overhead of these surrogate assisted MOEAs is usually significant for problems with a high-dimensional decision space, due to the 
high time complexity for  
the common techniques involved in these algorithms, such as population clustering, acquisition function optimization, and drop-out strategy. For example, in a test case with the ZDT test functions~\cite{Zitzler2000} of 30 decision variables, the data preprocessing before the GP model building in MOEA/D-EGO can  cost around 1 hour at the beginning, and the time complexity grew 
{cubically} with each iteration. 
ParEGO performed better with regard to the computation overhead. 
However, it suffers from a low sample  number per iteration, with only 1 sample data point per iteration, as compared to 5 or more in MOEA/D-EGO. Therefore, it is not time efficient and cannot take advantage of parallel 
computing capability, which is common in today's computing environments. 
In addition, the performance of the algorithms that employ the aforementioned techniques  suffer from the curse of dimensionality~\cite{curseofdim}. Therefore, they are rarely  used to solve problems with a relatively high  dimensional ($P > 20$) decision space~\cite{KRVEA}. 
The difficulty to extend to high-dimension problems and the high computation overhead for 
these algorithms limit their usefulness in many design studies.


{Neural network (NN)-based surrogate models have also been proposed to assist 
optimization algorithms~\cite{Syberfeldt2008,Liu2008,Kourakos2013}. 
The NN models are used to calculate or substitute the function evaluations, or help 
select an offspring solution for evaluation. 
}
A recent approach proposed in the accelerator field is to train a neural network as the surrogate model, from which the Pareto front 
can be found with optimization~\cite{EdelenML}. 
The challenge for this approach is that a large amount of data points may be needed to construct a global NN-based surrogate 
model that is sufficiently accurate to determine the Pareto front with it. 
Transfer learning, i.e., application of a learned model from one design problem to another,
is usually not applicable. This is because design problems, even for those with the same parameter setup, 
are very different from each other due to the nonlinear nature of the problems. 
Therefore, simulation data for one problem are typically of little use for another problem.


In this study, we propose a multi-objective multi-generation Gaussian process optimizer (MG-GPO) for design optimization. 
Similar to MOGA and MOPSO, it generates and manipulates a population of solutions with stochastic operations in an 
iterative manner. 
The difference is that posterior GP models are constructed and updated in each iteration and are used 
to select the trial solutions for the actual evaluation. 
The model-based selection substantially boosts the efficiency of the algorithm. 
The MG-GPO algorithm  differs from other surrogate-assisted evolutionary algorithms, such as MOEA/D-EGO and ParEGO, in that the GP models are only used for filtering, instead of for the generation 
of new solutions (e.g., through optimization). 
This reduces the requirement for high accuracy in the model 
{and consequently lowers the computation overhead, }
and give the algorithm high robustness and reliability. 
The implementation of the algorithm is also simple and straightforward. 
Because of such features, it is easy to apply to high-dimensional problems. 
{In addition, as non-dominated sorting is used both to select trial solutions with 
the GP models and to 
select the fittest solutions for the next generation, constraints can be easily 
integrated by including them in the sorting criteria. }


The paper is organized as follows. In Section~\ref{secGPintro} we give a brief introduction to the Gaussian 
process regression and optimization. 
The multi-generation GP optimizer is described in Section~\ref{secmgGPoptim}. 
A test of the new optimizer with analytic functions 
is presented in Section~\ref{secmgTestFunc}. The conclusion is given in Section~\ref{secConcl}.

\section{Gaussian process regression and optimization \label{secGPintro}}

The Gaussian process regression is a type of Bayesian inference, in which one combines 
a prior statistical model and the 
observed evidences to deduce knowledge of the actual statistical 
model, based on Bayes' theorem of the conditional probabilities.  

A Gaussian process is a statistical model of the distribution of a random function 
over space or time (distribution of a parameter space is assumed in the present context). 
The GP not only gives the probability distribution of the function 
at one location, but also its joint distribution with the function value at any other 
location. The joint distribution is a normal distribution. 
For an unknown function over a parameter space, a prior Gaussian process can be 
specified with the prior mean function $m({\bf x})$ and the kernel function 
$k({\bf x},{\bf x'})$, where ${\bf x}$ and ${\bf x'}$ are vectors that represent points 
in the parameter space. 
Without any knowledge about the function, the prior mean is often assumed $m({\bf x})=0$. 
The kernel function represents the covariance of the function values at two locations. 
It is often assumed to take the squared exponential form~\cite{RasmussenGP,  BrochuGPTut2010},
\begin{eqnarray}
    k({\bf x},{\bf x'}) = \Sigma_f^2 \exp(-\frac12 ({\bf x}-{\bf x'})^T{\bf \Theta}^{-2}({\bf x}-{\bf x'})),
\end{eqnarray}
where $\Sigma_f$ is the estimated variance of the function,
${\bf \Theta}=\text{diag}(\theta_1, \theta_2, \cdots, \theta_n)$ is a diagonal 
matrix and the $\theta_i$ parameters specify the correlation of the function values 
at two points separated in space in the direction of $x_i$ coordinate. 

After a number of sample data points,  
given as $({\bf x}_i, f_i=f({\bf x}_i))$, 
$i=1$, $2$, $\cdots$, $t$,  are taken from the parameter space, 
we would like to know the function value at a new point ${\bf x}_{t+1}$. 
From the prior GP, the joint distribution of the sample data and the new point is given 
by a multi-variate normal distribution, 
\begin{eqnarray}\label{eq:priorGPdistr}
     \mathcal{N}\left({\bf 0}, \begin{pmatrix} 
    {\bf K} & {\bf k} \\ {\bf k}^T & k({\bf x}_{t+1}, {\bf x}_{t+1}),
    \end{pmatrix} \right)
\end{eqnarray}
where ${\bf K}$ is the kernel matrix, whose elements are $K_{ij}=k({\bf x}_i,{\bf x}_j)$, 
and the kernel vector is given by $k_i=k({\bf x}_i,{\bf x}_{t+1})$. 
The prior joint distribution function, Eq.~\ref{eq:priorGPdistr}, and the evidence by 
the sample data set allow us to calculate the conditional distribution of the 
function value at point  ${\bf x}_{t+1}$, which is a normal distribution given by 
its mean and standard deviation~\cite{RasmussenGP},
\begin{eqnarray}
      \mu_{t+1} &=& {\bf k}^T{\bf K}^{-1}{\bf f}_t,  \label{eq:postMean} \\
    \sigma^2_{t+1} &=& k({\bf x}_{t+1}, {\bf x}_{t+1})-{\bf k}^T{\bf K}^{-1}{\bf k}. \label{eq:postSig}
\end{eqnarray}
The expected mean, ${\mu}_{t+1}$, is an estimate of the function value and 
the standard deviation $\sigma_{t+1}$ gives the uncertainty. 

Eqs.~\ref{eq:postMean}-\ref{eq:postSig} are the posterior model of the 
actual function. It is worth noting that this is a non-parametric model. 
The sample data enter the model directly. 
The posterior distribution not only can be used to predict
the function values in the parameter space, but also can be used to optimize the function. 

In a GP optimizer, the posterior model is used to choose the next trial solution. 
With the posterior GP, an optimization algorithm is used to look for a point ${\bf x}_{t+1}$ that 
the model predicts to yield the largest gain, which is then evaluated on the real 
system. After that, the new data point enters the sample data set and the GP model is 
updated accordingly. The measure of the gain is represented by the acquisition function, 
a popular choice of which is the upper confidence bound (UCB) for 
a maximization problem~\cite{GPUCB}. For a minimization problem, it is the lower confidence bound (LCB), 
given by 
\begin{eqnarray}
    \text{GP-LCB}({\bf x}) = \mu({\bf x})-\kappa \sigma({\bf x}),
\end{eqnarray}
where $\kappa\ge 0$ is a constant. A suitable value of $\kappa$ is used to balance the 
exploitation 
and the exploration strategies - a small $\kappa$ favors exploitation and a large $\kappa$ 
favors exploration. 
Taking a large $\kappa$ is to take some risk by going into the less certain area in the parameter space 
in exchange for the opportunity to yield a big gain. 

After every new data point is added, the GP model is updated, which requires the inversion of the 
kernel matrix. During the search for the trial solution, many matrix multiplications are 
performed. These calculation can be time consuming if the dimension of the matrix is large. 
Therefore, the size of the data set is often limited to the order of hundreds. 

\section{Multi-generation Gaussian process optimizer \label{secmgGPoptim}}
The ability of the posterior GP model to approximate the actual model and to predict the performance 
of a new solution can  be very useful in design optimization, where it is common to evaluate 
thousands or tens of thousands solutions in the search for the optimal design. 
A design study often has multiple objectives. 
In the following we propose a multi-objective, multi-generation Gaussian process optimization 
algorithm that would be ideal for design optimization. 

Presently MOGA and MOPSO algorithms are widely used in the design optimization of accelerators. 
A popular MOGA algorithm is the NSGA-II~\cite{NSGA2}. 
It takes an iterative scheme to update a population of solutions. At each iteration, it generates 
new trial solutions based on the existing ones, using the 
simulated binary crossover (SBX)~\cite{DebSBX1995} and polynomial mutation operations~\cite{liagkouras2013elitist}.
In a crossover two solutions are combined to generate a pair of new solutions randomly 
distributed in between, while a mutation operation modifies a solution with random changes to the 
parameters. The new trial solutions are evaluated and compared to the existing solutions with a 
non-dominated sorting. Some solutions replace the existing ones and enter the next generation if 
they outperform the latter. 

The MOPSO~\cite{Kennedy488968,PSO985692Kennedy} 
algorithm also manipulates a population of solutions iteratively. In this case, each solution is 
considered a particle in the parameter space. New solutions 
are generated by shifting the existing solutions in the parameter space by an offset called the velocity. 
The velocity consists of contributions from three terms: 
the previous velocity, a shift toward the best solution of the history of the particle 
(the personal best), and a shift 
toward a solution in the global best solutions. The velocity and the personal and global best solutions 
are updated at every iteration. 

The MOGA and MOPSO algorithms work because the operations used to generate new solutions tend to 
produce solutions toward the direction with better performances, which are then selected and used
for the next generation. 
However, there is no guarantee that the crossover and mutation operations
or the shift by the velocity will 
yield better solutions. 
No information is extracted from the previous function evaluations other than the selection of the 
best solutions. 

When we apply  GP regression to model the existing solutions, 
we would be able to determine which new
solutions have a high probability of yielding good performances. 
We can optimize with the posterior GP model to produce promising trial solutions. 
Or we can simply generate a large quantity of potential new solutions,  evaluate them with 
the GP model, and use the outcome to select the solutions with a potential to yield a significant 
improvement. 
By selecting only these 
solutions for the computationally expensive function evaluation, we could substantially improve the 
efficiency of the algorithm. 

The new algorithm, which may be referred to as the multi-objective,
multi-generation Gaussian process optimizer (MG-GPO), 
also works iteratively. 
The initial population of solutions may be randomly generated, throughout the parameter space, or 
within a small region in the parameter space. The population of solutions, $N$, is fixed.

At each iteration, $N$ new solutions will be generated and evaluated. The set of solutions evaluated 
on iteration $n$ may be labeled ${\mathcal F}_n$.  The set ${\mathcal F}_n$ is 
combined with the $N$  best solutions from the last iteration, 
which form a set labeled ${\mathcal G}_{n-1}$, and the 
combined set is sorted with the non-dominated sorting~\cite{NSGA2}, from which the population of $N$ best solutions 
is updated. 

A GP model is constructed for each objective, which 
has its own set of model parameters, ${\bf \Theta}^{(j)}$ and $\Sigma_f^{(j)}$. 
We also give the prior GP model a non-zero mean, $m_j({\bf x})=\bar{\mu}^{(j)}$. The value of $\bar{\mu}^{(j)}$ and 
$\Sigma_f^{(j)}$ are given by the mean and standard deviation of the 
function values of the previous data set, respectively. 
With the non-zero mean, Eq.~\ref{eq:postMean} is replaced with 
\begin{eqnarray}
          \mu_{t+1} &= {\bf k}^T{\bf K}^{-1}({\bf f}_t-\bar{\mu})+\bar{\mu}.  \label{eq:postMeanN}
\end{eqnarray}
The use of a non-zero mean helps avoid an abrupt change in the function value when searching 
in the transition region 
between the sampled area and the un-sampled areas. A wrong mean value 
could produce a bias that either pull the search into the unexplored territory or prevent the 
search into it. 

While it is possible to use a multi-objective optimization algorithm 
to optimize the surrogate models,  produce the Pareto front,  
and use the solutions in the Pareto front for the actual evaluation, 
we adopt a simple approach to sample the area around 
the existing best solutions. 
New solutions are generated through the mutation and crossover operations. 
For each solution in the previous population of best solutions, ${\mathcal G}_{n-1}$, 
$m_1$
new solutions are by mutation and another $m_2$
solutions are by crossover. 
Mutation is done by the 
{polynomial mutation (PLM) technique~\cite{liagkouras2013elitist}}.
Crossover is done with the SBX technique, as is done in NSGA-II.
Obviously, there could be better ways to generate new solutions, for example, by using the gradient 
afforded by the posterior GP model. 
Nonetheless, the present simple approach is adequate to demonstrate the advantage of the GP method. 
Besides, we can always increase the number of new solutions to improve the sampling of the 
GP models as the cost of evaluating the GP models is usually negligible compared to the 
actual physics simulation. 

The $(m_1+m_2)N$ solutions are then evaluated with the GP 
models, which give the expected mean and standard deviation 
for each objective function. 
We choose the GP-LCB acquisition functions as the figure of merit for the solutions. 
A non-dominated sorting is then performed over the $(m_1+m_2)N$ solutions, 
from which $N$ solutions are 
selected for the actual design simulation. 
These $N$ solutions form the set ${\mathcal F}_{n}$, which
is then combined with ${\mathcal G}_{n-1}$ and 
another non-dominated sorting is used to updated the  $N$ best solutions, 
yielding ${\mathcal G}_{n}$. 

\begin{figure}[!t]
\centering
\includegraphics[width=0.8\linewidth]{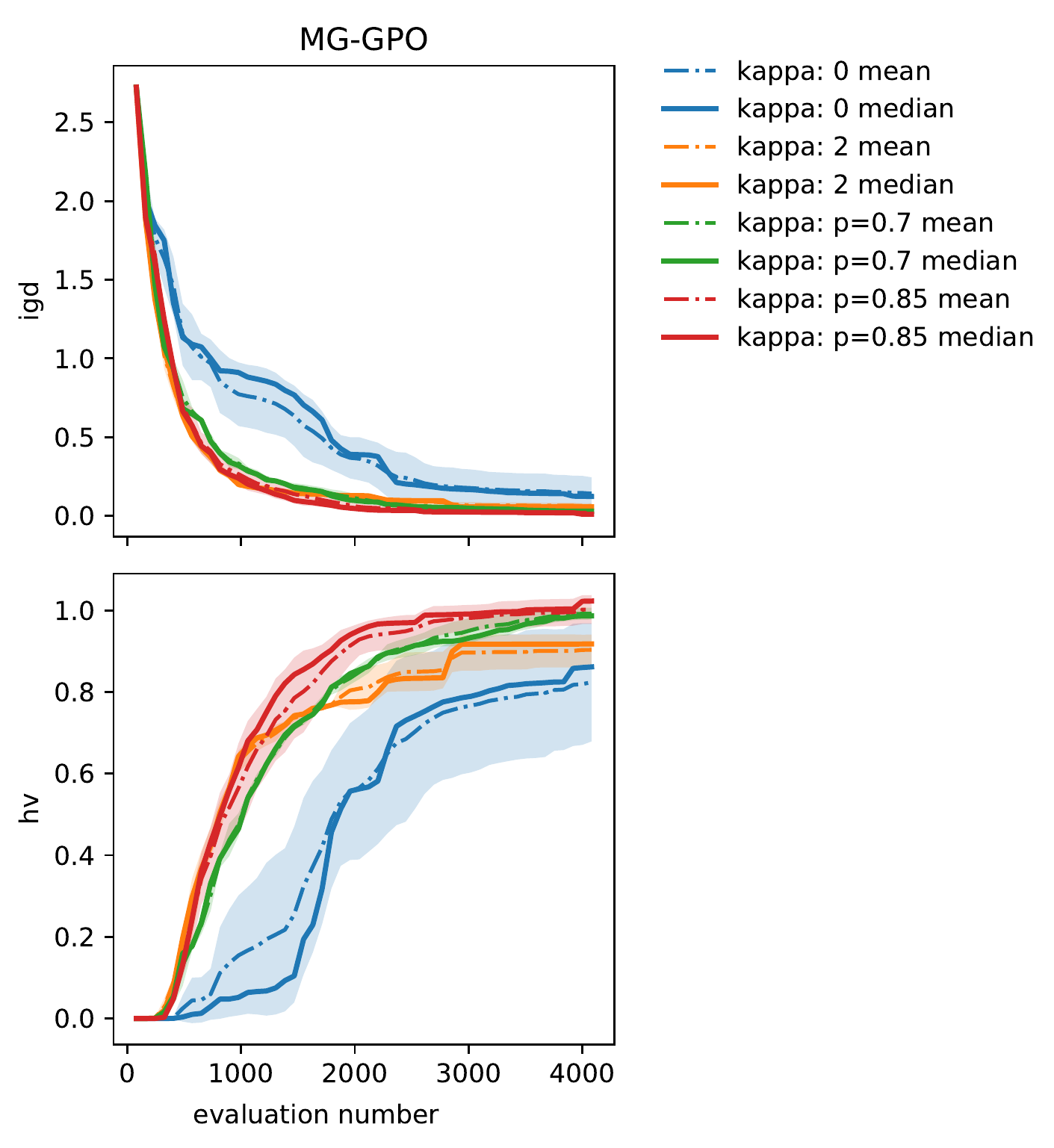}
\caption{The IGD and HV evolutionary traces of MG-GPO 
for the ZDT3 test problem with various $\kappa$ schemes for the GP-LCB acquisition function, 
including constant $\kappa=0$, $\kappa=2$, and an exponentially decaying $\kappa$
according to 
$\kappa_{n+1}=\rho \kappa_n$ from the initial value of 2, and $\rho=0.70$ and $0.85$. 
The mean and median curves are shown for 10 test runs for each case. The shaded areas indicate the 
one-sigma spread. The reference point in the HV calculation is set to $(1, 1)$.
}
\label{fig:perf_kappa}
\end{figure}
\begin{figure*}[htbp]
\centering
\includegraphics[width=0.95\linewidth]{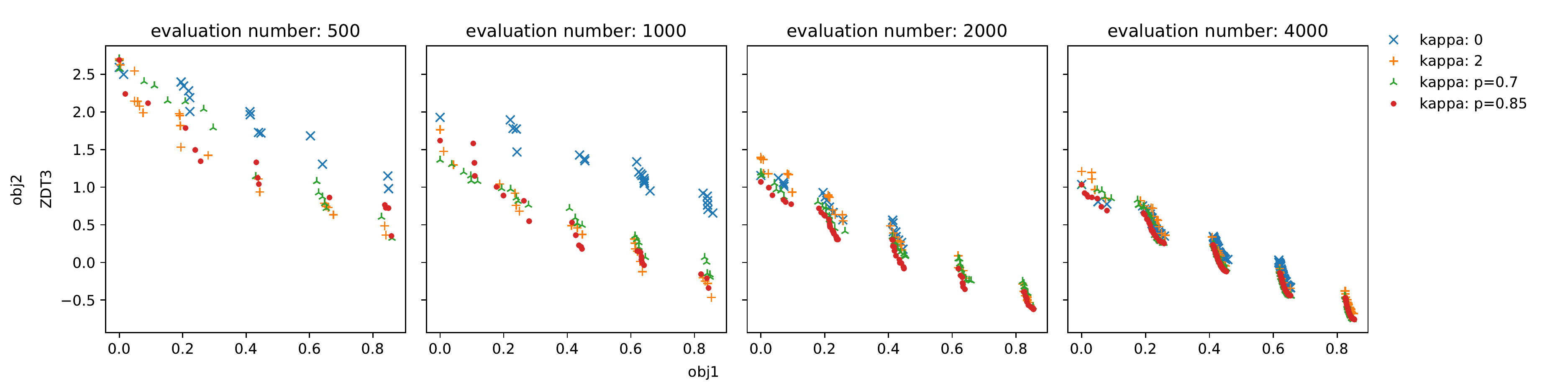}
\caption{Pareto fronts at specific evaluation numbers of MG-GPO with different $\kappa$ decay rates for the ZDT3 test problem. The case with the median performance is shown for each 
test case. }
\label{fig:front_kappa}
\end{figure*}

The $\kappa$ parameter in GP-LCB can have a significant impact to the behavior of the algorithm. 
A large $\kappa$ value encourages exploration of the parameter space but in the same time may not 
take full advantage of the learned model. 
Conversely, a small $\kappa$ value better exploits the model  but may not sufficiently explore 
the parameter space. 
It could be argued that at the beginning of an optimization a large $\kappa$ is preferred as more 
exploration is needed in order to discover the area in the parameter space with good solutions. 
A small $\kappa$ would be preferred in the later stage as the algorithm converges to a relatively 
small area where a refined search is needed. 
Figure~\ref{fig:perf_kappa} and \ref{fig:front_kappa} compares the convergence of the MG-GPO algorithm with several $\kappa$ 
schemes for the ZDT3 test case (see next section). It was found that decreasing $\kappa$  
exponentially generation by generation from $\kappa=2$ toward zero, with 
$\kappa_{n+1}=\rho \kappa_n$, $\rho=0.85$ gives the best performance for the problem. 
The optimal decreasing factor $\rho$ may depend on the optimization problem and the optimization setup. 
Ideally, an 
adaptive scheme of decreasing $\kappa$ over generations, using certain performance metrics (e.g., 
hyper-volume) as the guide would be preferred. 


The GP models are updated at the end of the iteration. 
The sample data used for the GP models are the combined set of ${\mathcal F}_{n}$ and ${\mathcal G}_{n}$. 
There will be some redundant data points, as some solutions in ${\mathcal F}_{n}$ has just entered 
${\mathcal G}_{n}$. The duplicate points can be eliminated. 
It can also be left in, as it does not pose a difficulty, 
{when singular value decomposition (SVD) is used in the 
matrix inversion in Eq.~\ref{eq:postMeanN}.}. 

The MG-GPO algorithm  is summarized below (with $G_{max}$ being the maximum number of 
generations, $\rho$  the decreasing factor for $\kappa$)
\begin{algorithmic}
\STATE $n\gets 0$, Initialize the population, $\mathcal{G}_0$. 
{Initialize $\kappa \gets 2$. }
\STATE Evaluate all solutions in $\mathcal{G}_0$
\STATE Construct Gaussian process models, $\mathcal{GP}_0$, with $\mathcal{G}_0$
\WHILE {$n < G_{max}$} 
        \STATE $n\gets n+1$
        \STATE {Update $\kappa$ with $\kappa \gets \rho \kappa$. }
        \STATE For each solution in $\mathcal{G}_{n-1}$, 
        generate $m_1$ solutions with mutation and $m_2$ solutions with crossover.
        \STATE Evaluate the $(m_1+m_2)N$ solutions with $\mathcal{GP}_{n-1}$
        \STATE Use non-dominated sorting to select $N$ best solutions, which forms the 
        set $\mathcal{F}_n$.
        \STATE Evaluate the solutions in  $\mathcal{F}_n$ in the actual system. 
        \STATE Use non-dominated sorting to select $N$ best solutions from the 
        combined set of $\mathcal{G}_{n-1}$ and  $\mathcal{F}_n$, the results of which form  $\mathcal{G}_{n}$.
        \STATE Construct Gaussian process models, $\mathcal{GP}_n$, with solutions in 
         $\mathcal{F}_n$ and $\mathcal{G}_n$. 
\ENDWHILE
\end{algorithmic}

{In the above algorithm, new trial solutions are generated with mutation and crossover 
operations, as is done in NSGA-II. The new trial solutions can also been generated with the 
moving particle approach as done in MOPSO. The attraction terms by the global and personal best 
solutions can be varied and selected by the GP models. We tested the approach and found that 
similar improvement in convergence efficiency can be made. }

\section{Performance comparisons with other algorithms\label{secmgTestFunc}}

Simulations were conducted to demonstrate the fast convergence of the MG-GPO algorithm in comparison with a few commonly used multi-objective optimization algorithms. 
The  algorithms selected for comparison include two classic algorithms, NSGA-II and MOPSO, 
and two more recent ones, MMOPSO~\cite{MMOPSO} and WOF-SMPSO~\cite{WOFSMPSO2016,WOFSMPSO2017,WOFSMPSO2018}.
The NSGA-II, MMOPSO, and WOF-SMPSO codes used in the tests are from the PlatEMO platform\cite{PlatEMO}. The MOPSO algorithm is the same as used in Ref.~\cite{HUANG201448PSO} and 
was  based on the framework described in~\cite{PANG2014124PSO}.


\subsection{Test Instances}

Four test cases have been used to test the performance of the MG-GPO algorithm in 
comparison to the NSGA-II, MOPSO, MMOPSO and WOF-SMPSO algorithms. 
These test cases are commonly used for algorithm performance comparison, for example,
in Ref.~\cite{NSGA2}. All test cases have two objective functions, and assumed to be minimization problems. 

\subsubsection{ZDT1~\cite{Zitzler2000}}

The ZDT1 test case is defined by
\begin{eqnarray}
    f_1({\bf x})&=&x_1,\\
    f_2({\bf x})&=&g({\bf x})(1-\sqrt{x_1/g({\bf x})}),
\end{eqnarray}
with
\begin{eqnarray}
    g({\bf x})=1+\frac{9}{P-1}(\sum_{i=2}^P x_i).
\end{eqnarray}
The parameter ranges are [0, 1] for all variables. 
The optimal solutions are with $x_i=0$ for $i=2$, 3, $\cdots$, $P$ and 
$x_1\in [0, 1]$. 


\subsubsection{ZDT2~\cite{Zitzler2000}}
ZDT2 is defined similarly as ZDT1, except that the $f_2({\bf x})$ function is redefined as
\begin{eqnarray}
    f_2({\bf x})=g({\bf x})\left[1-({x_1/g({\bf x}))^2}\right]. 
\end{eqnarray}
The ranges of the parameters are the same as ZDT1.

  
\subsubsection{ZDT3~\cite{Zitzler2000}}
The definition is similar to ZDT1, except that the $f_2({\bf x})$ function is redefined as
\begin{eqnarray}
    f_2({\bf x})=g({\bf x})(1-\sqrt{x_1/g({\bf x})}-\frac{x_1}{g({\bf x})} \sin10\pi x_1 ). 
\end{eqnarray}
The ranges of the parameters are the same 
as ZDT1. The Pareto front of this case consists of disconnected stripes. 


\subsubsection{ZDT6~\cite{Zitzler2000}}

ZDT6 test case is defined as
\begin{eqnarray}
    f_1({\bf x})&=&1-e^{-4x_1}\cdot\sin^6(6\pi x_1),\\
    f_2({\bf x})&=&1-\left(\frac{f_1({\bf x})}{g({\bf x})}\right)^2,
\end{eqnarray}
with
\begin{eqnarray}
    g({\bf x})=1+9\left[\sum_{i=2}^P x_i / (P-1)\right]^{0.25}.
\end{eqnarray}
The ranges of the parameters are the same 
as ZDT1. Its Pareto front is nonconvex, and the distribution of the Pareto solutions is highly nonuniform.

The dimension of the test cases was chosen to be $P=30$. 
This may represent the mid-dimensional decision spaces in design studies. 
Tests were also done with $P=100$ for  ZDT1 and ZDT2.
The $P=100$ cases represent high-dimensional design problems. 


\subsection{Experimental Settings}

\subsubsection{General Settings}

For the MG-GPO implementation, the parameter range is normalized to $[0, 1]$. 
The mutation (PLM) and crossover (SBX) control parameters are set to $\eta_m=20$ and $\eta_c=20$, respectively. 
{To balance the need for initial exploration in the early stage and 
refined search in the later stage, the $\kappa$ parameter in the GP-LCB acquisition function is 
exponentially decreased,  initially with $\kappa=2$ and with each generation it is scaled down by 
the factor $\rho=0.85$. 
}
The multiplication factors are set to $m_1=m_2=20$. 
In all test cases, the correlation length parameters of MG-GPO are optimized in each generation with 
respect to 
the  data samples used for GP model construction.
{The GPy package is used for hyper-parameter optimization~\cite{gpy2014}}.

For NSGA-II, the crossover probability is set to $90\%$. 
The distribution indices for the simulated binary crossover (SBX) and mutation operations 
are $\eta_c=20$ and $\eta_m=20$, respectively~\cite{DebSBX1995}.

For MOPSO, the weight factors in the velocity composition are 
$w=0.4$ and $r_1=r_2=1$. The MOPSO algorithm also includes a mutation operation, with a 
probability rate of $1/P$, where $P$ is the number of variables. 

For MMOPSO, there are no algorithm-specific hyper parameters to modify.

For WOF-SMPSO, all hyper parameters are set to the default values, as suggested in Ref.~\cite{WOFSMPSO2018}.

The initial solutions are randomly distributed, with parameters drawn 
from a uniform distribution in the parameter range. 

We chose Hypervolume (HV)~\cite{zitzler2003performance} and inverted generational distance (IGD)~\cite{coello2004study, sierra2004new} as optimization algorithm performance indicators (PI).

\subsubsection{Mid-dimensional Settings}

The mid-dimensional test problems are ZDT1, ZDT2, ZDT3 and ZDT6.
The  dimensions of the test problems are set to $P=30$.
The population size is set to $N=80$ for all algorithms. 
The algorithms are run for 4080 evaluations. Each test instance was repeated 10 times.

\subsubsection{High-dimensional Settings}

The high-dimensional test problems are ZDT1 and ZDT2.
The dimensions of the test problems are set to $P=100$.
The population size is set to $N=80$ for all algorithms. 
The algorithms are run for 8080 evaluations. Each test instance was repeated 10 times.

\subsection{Experimental Results}

\begin{figure*}[htbp]
\centering
\includegraphics[width=0.95\linewidth]{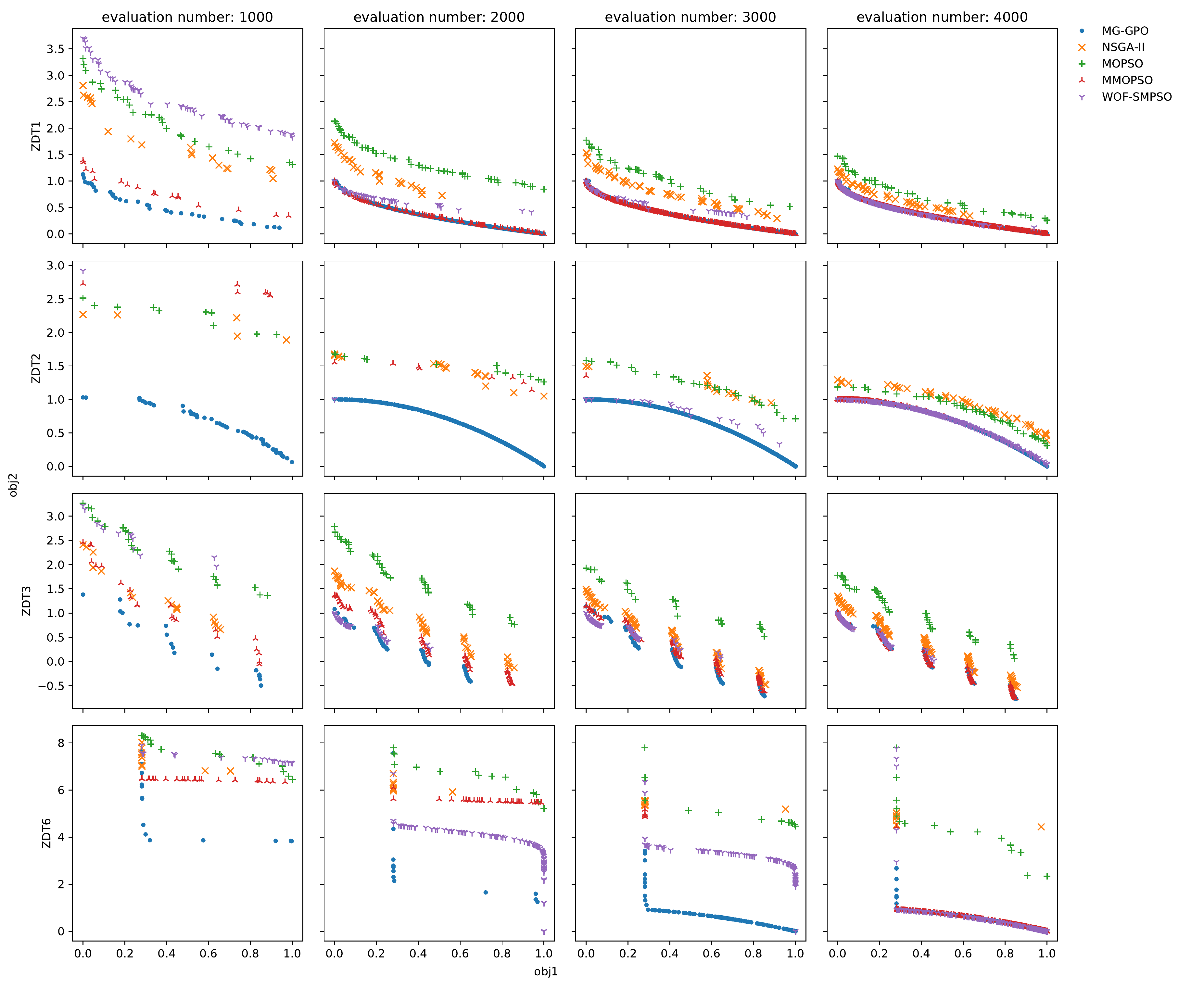}
\caption{The dominant front in the population of solutions
 at four instants during the optimization runs for the 
algorithms being compared, MG-GPO, NSGA-II, MOPSO, MMOPSO and WOF-SMPSO, for the $P=30$ cases (mid-dimensional setting). 
The run whose performance ranks the fifth among the 10 runs for each algorithm is shown.  
Rows from top to bottom are for test problems ZDT1, ZDT2, ZDT3, and ZDT6, respectively. 
The columns, from left to right, are for the front at evaluation number 1000, 2000, 3000, 
and 4000, respectively.}
\label{fig:front30}
\end{figure*}

Figure~\ref{fig:front30} shows four snapshots of 
the distribution of the non-dominated front in the population of solutions during the course 
of optimization for the five algorithms being compared for all four test problems 
with $P=30$, with the number of evaluations from 1000 to 4000, at an 1000 interval. 
The case with the median behavior among the 10 tests for each algorithm is shown.

For all four test problems, MG-GPO converges the fastest among the five algorithms. 
For ZDT1, MG-GPO nearly converges to the Pareto front within 1000 evaluations, followed by 
MMOPSO and WOF-SMPSO.
For ZDT2, MG-GPO also converges within approximately 1000 evaluations, followed by WOF-SMPSO, which 
converges with 4000 evaluations. 
For ZDT3 and ZDT6, MG-GPO takes 2000 and 3000 evaluations to converge, respectively. 

\begin{figure*}[htbp]
\centering
\includegraphics[width=\linewidth]{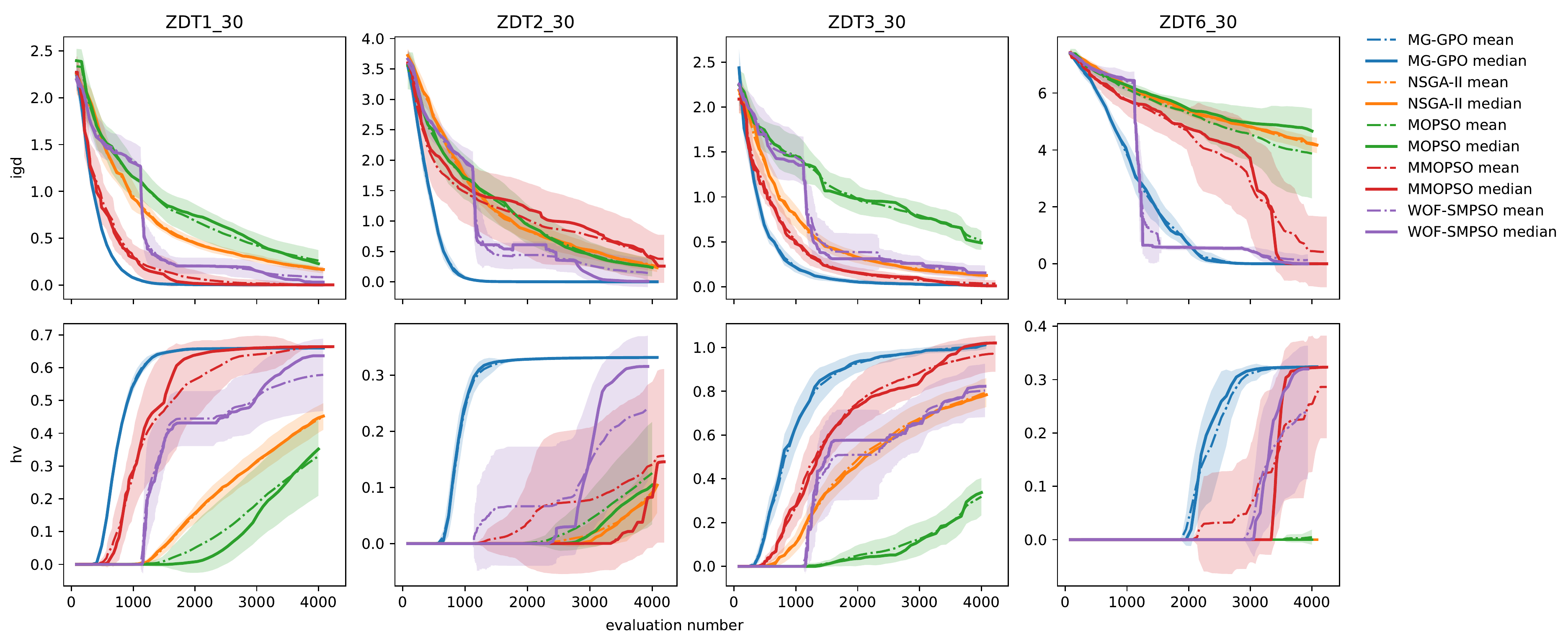}
\caption{Comparison of HV and IGD between the five algorithms in the mid-dimensional settings ($P=30$) experiments. 
Top row for  IDG  and the bottom row for HV. 
The columns stand for four test problems. The filled area around each mean curve indicates the standard deviation of the performance metric for each algorithm. The reference point in the HV calculation is set to $(1, 1)$.}
\label{fig:perf30}
\end{figure*}

To better describe the convergence history of the algorithms, Figure~\ref{fig:perf30} 
shows the evolution of HV and IGD metrics for the algorithms for the $P=30$ cases. 
For each algorithm, the median and the mean values for the 10 tests are shown. The spread of the 
metrics among the 10 tests for each algorithm is shown with shaded areas. 
Clearly, MG-GPO converges faster than the other algorithms. 
In addition, the performance of MG-GPO is very stable. The spread of the metrics for MG-GPO is 
considerably smaller than the other algorithms.





\begin{figure*}[htbp]
\centering
\includegraphics[width=0.95\linewidth]{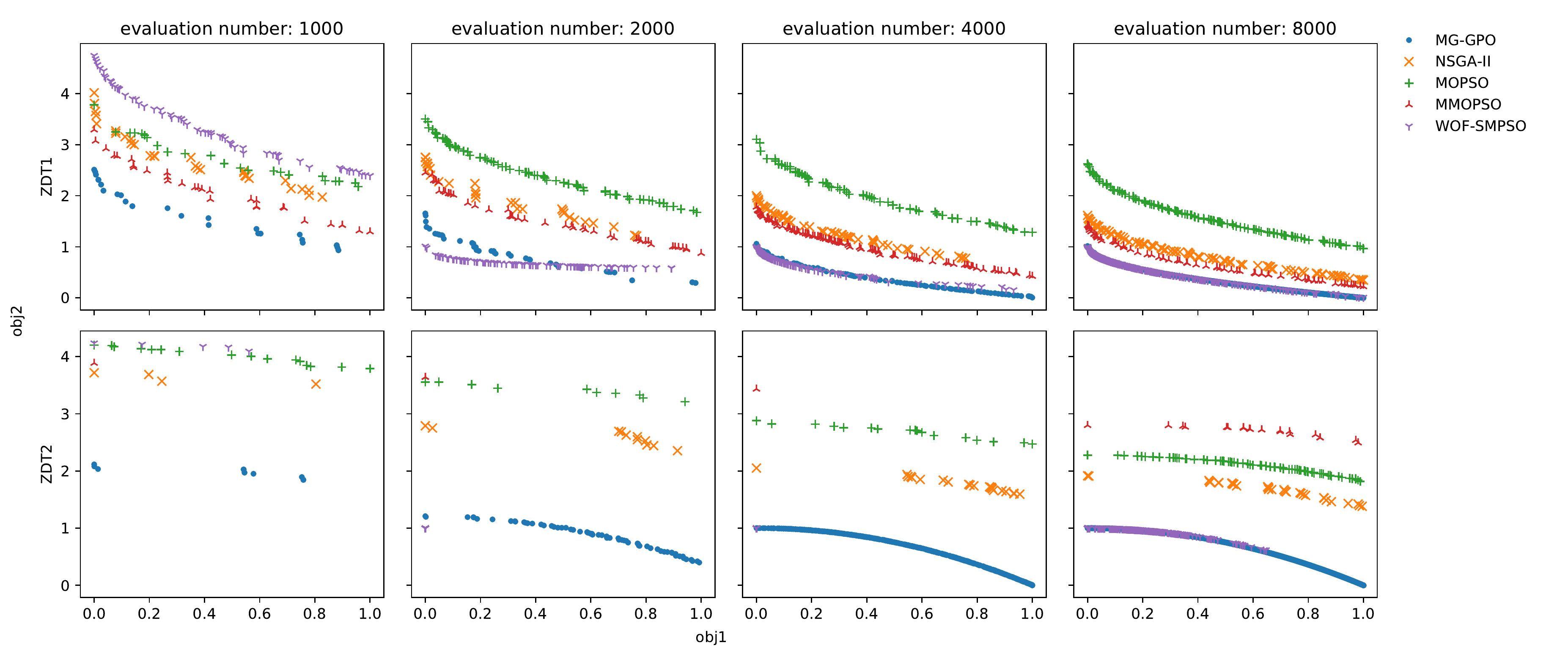}
\caption{
{The dominant front in the population of solutions
 at four instants during the optimization runs for the 
algorithms being compared, MG-GPO, NSGA-II, MOPSO, MMOPSO and WOF-SMPSO, for the $P=100$ case (high-dimensional setting). 
The run with the fifth performance among the 10 runs for each algorithm is shown.  
Top row for test problem ZDT1 and bottom row for ZDT2. 
The columns, from left to right, are for the front at evaluation number 1000, 2000, 4000, 
and 8000, respectively.}}
\label{fig:front100}
\end{figure*}
\begin{figure}[htbp]
\centering
\includegraphics[width=1.1\linewidth]{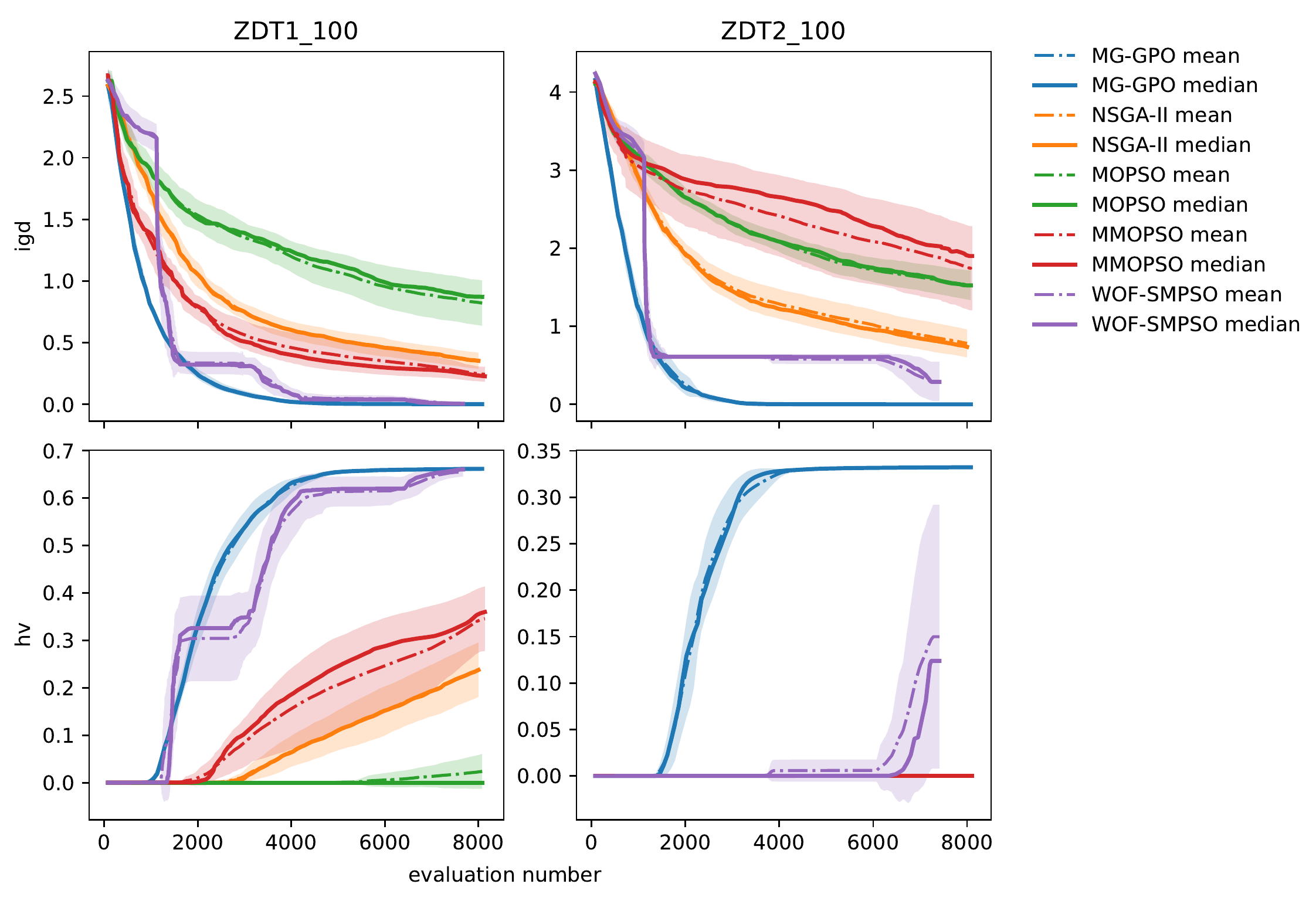}
\caption{Comparison of HV and IGD between different algorithms in the high-dimensional ($P=100$) experiments. 
Top row for IGD and bottom row for HV. 
Left column for ZDT1 and right column for ZDT2. The reference point in the HV calculation is set to $(1, 1)$.
}
\label{fig:perf100}
\end{figure}

Simulations for test cases ZDT1 and ZDT2 with $P=100$ were done to demonstrate the performance of the 
MG-GPO algorithm for high-dimensional optimization problems. 
Figure~\ref{fig:front100} shows the non-dominated front of solutions for the algorithms at four 
generations. For both problems, MG-GPO converges to the Pareto front within 4000 evaluations, leading 
the other algorithms by a substantial margin. 
The evolution of HV and IGD metrics is shown in Figure~\ref{fig:perf100}, which clearly indicates that 
MG-GPO does not only converge fastest, but also has a stable performance.  

{The algorithm performance comparison results are also summarized in 
tables. Table~\ref{Table:IGD30} and \ref{Table:HV30} show the 
IGD and HV metrics, respectively, for the $P=30$ cases. 
The best value, the means, and the standard deviations (for the 10 runs) are listed for 
each algorithm at the four instants during the runs. 
Similarly,  
Table~\ref{Table:IGD100} and \ref{Table:HV100} show the results for the 
$P=100$ cases. 
The tables show that MG-GPO converges significantly faster than the other algorithms. }

{
The results of the Wilcoxon tests at a 0.05 significance level of MG-GPO vs. the other four algorithms are listed in 
Table~\ref{table:wilcoxon_hv_30D} and \ref{table:wilcoxon_igd_30D} for the HV and IGD metrics,
respectively, for the $P=30$ cases. Table~\ref{table:wilcoxon_hv_100D} and \ref{table:wilcoxon_igd_100D} show the results for the $P=100$ cases. In the tables, the value 1 denotes that MG-GPO wins, 0 for no winner, -1 for that the other algorithm wins, and N/As in the HV tables denote that neither algorithm has reached a positive HV among the 10 runs. For the $P=30$ 
IGD tests, MG-GPO wins all tests against NSGA-II and MOPSO, wins 11 times and 
draws 5 times against MMOPSO, and wins 15 times and draws 1 time against WOF-SMPSO (for a total of 
16 tests between the two algorithms). Similar result were found for the $P=30$ HV tests. 
Note that MG-GPO loses once to MMOPSO on ZDT1 at evaluation number 4000, but cross examination with Table~\ref{table:compare_hv_all_30D} shows that the mean HV difference is quite small ($< 0.4\%$). 
For the $P=100$ cases, MG-GPO  wins almost all the tests, except with 1 draw with WOF-SMPSO on ZDT1 at evaluation number 2000 in IGD and HV.
The Wilcoxon test results clearly indicate that MG-GPO has a significant better performance over the other algorithms being compared. This performance advantage is even bigger for the higher dimensional problems.}

\begin{table*}[htbp]
\centering
\caption{ Comparison of IGD for the algorithms for the $P=30$ cases. The best results are highlighted.
\label{Table:IGD30} }
\begin{tabularx}{0.65\linewidth}{l l l r r r r r}
\toprule
Eval \# & Instance & & MG-GPO & NSGA-II & MOPSO & MMOPSO & WOF-SMPSO \\
\midrule
1000 & ZDT1\_30 & best &  {\bf 0.0514} &  0.7387 &  0.7346 &  0.1476 &    0.9550 \\
     &         & mean &  {\bf 0.0759} &  0.9481 &  1.1539 &  0.3047 &    1.3268 \\
     &         & std &  {\bf 0.0187} &  0.1595 &  0.2072 &  0.1392 &    0.2066 \\
     & ZDT2\_30 & best &  {\bf 0.0489} &  1.4296 &  1.0101 &  0.6318 &    1.5738 \\
     &         & mean &  {\bf 0.0755} &  1.7034 &  1.7217 &  1.4759 &    1.9935 \\
     &         & std &  {\bf 0.0305} &  0.1424 &  0.3852 &  0.5437 &    0.2289 \\
     & ZDT3\_30 & best &  {\bf 0.1556} &  0.5776 &  1.1358 &  0.3602 &    1.1512 \\
     &         & mean &  {\bf 0.2206} &  0.7802 &  1.4615 &  0.5089 &    1.4399 \\
     &         & std &  {\bf 0.0653} &  0.1334 &  0.2403 &  0.1268 &    0.2780 \\
     & ZDT6\_30 & best &  {\bf 2.8674} &  6.1099 &  5.6077 &  4.5536 &    5.5248 \\
     &         & mean &  {\bf 3.8390} &  6.2376 &  6.1368 &  5.7419 &    6.4118 \\
     &         & std &  0.5359 &  {\bf 0.0845} &  0.2968 &  0.5492 &    0.3631 \\
2000 & ZDT1\_30 & best &  {\bf 0.0033} &  0.3784 &  0.3477 &  0.0049 &    0.0509 \\
     &         & mean &  {\bf 0.0050} &  0.4532 &  0.6877 &  0.0728 &    0.2064 \\
     &         & std &  {\bf 0.0015} &  0.0535 &  0.1887 &  0.0916 &    0.0914 \\
     & ZDT2\_30 & best &  {\bf 0.0021} &  0.6264 &  0.4597 &  0.0444 &    0.0062 \\
     &         & mean &  {\bf 0.0028} &  0.8399 &  0.9464 &  1.0236 &    0.4403 \\
     &         & std &  {\bf 0.0006} &  0.1596 &  0.3391 &  0.6377 &    0.2350 \\
     & ZDT3\_30 & best &  {\bf 0.0295} &  0.2351 &  0.6616 &  0.0358 &    0.1530 \\
     &         & mean &  {\bf 0.0586} &  0.3270 &  0.9632 &  0.1489 &    0.3860 \\
     &         & std &  {\bf 0.0274} &  0.0651 &  0.1657 &  0.0685 &    0.2171 \\
     & ZDT6\_30 & best &  {\bf 0.2157} &  4.9799 &  4.3317 &  2.6347 &    0.4318 \\
     &         & mean &  0.6519 &  5.3506 &  5.2870 &  4.6126 &    {\bf 0.5668} \\
     &         & std &  0.3303 &  0.1623 &  0.5075 &  0.8643 &    {\bf 0.0888} \\
3000 & ZDT1\_30 & best &  0.0023 &  0.1982 &  0.1664 &  {\bf 0.0015} &    0.0138 \\
     &         & mean &  {\bf 0.0033} &  0.2827 &  0.4234 &  0.0178 &    0.1485 \\
     &         & std &  {\bf 0.0006} &  0.0427 &  0.1387 &  0.0429 &    0.0975 \\
     & ZDT2\_30 & best &  {\bf 0.0010} &  0.3120 &  0.1888 &  0.0020 &    0.0038 \\
     &         & mean &  {\bf 0.0012} &  0.5040 &  0.4565 &  0.7575 &    0.2707 \\
     &         & std &  {\bf 0.0002} &  0.1360 &  0.2140 &  0.5398 &    0.2674 \\
     & ZDT3\_30 & best &  0.0135 &  0.1420 &  0.5264 &  {\bf 0.0082} &    0.1341 \\
     &         & mean &  {\bf 0.0318} &  0.1992 &  0.7939 &  0.0764 &    0.2507 \\
     &         & std &  {\bf 0.0151} &  0.0492 &  0.1538 &  0.0505 &    0.0974 \\
     & ZDT6\_30 & best &  0.0033 &  4.3218 &  2.0640 &  {\bf 0.0021} &    0.3508 \\
     &         & mean &  {\bf 0.0118} &  4.7891 &  4.5506 &  2.8612 &    0.4954 \\
     &         & std &  {\bf 0.0118} &  0.2426 &  1.0071 &  1.8952 &    0.1161 \\
4000 & ZDT1\_30 & best &  0.0019 &  0.1060 &  0.0784 &  {\bf 0.0008} &    0.0056 \\
     &         & mean &  0.0029 &  0.1655 &  0.2603 &  {\bf 0.0023} &    0.0845 \\
     &         & std &  {\bf 0.0006} &  0.0363 &  0.1192 &  0.0020 &    0.1117 \\
     & ZDT2\_30 & best &  {\bf 0.0007} &  0.1629 &  0.0602 &  0.0008 &    0.0008 \\
     &         & mean &  {\bf 0.0008} &  0.2782 &  0.2364 &  0.4185 &    0.1452 \\
     &         & std &  {\bf 0.0001} &  0.1045 &  0.1550 &  0.4182 &    0.2458 \\
     & ZDT3\_30 & best &  0.0067 &  0.0803 &  0.3928 &  {\bf 0.0040} &    0.0454 \\
     &         & mean &  {\bf 0.0205} &  0.1339 &  0.5158 &  0.0372 &    0.1578 \\
     &         & std &  {\bf 0.0173} &  0.0406 &  0.1102 &  0.0425 &    0.0854 \\
     & ZDT6\_30 & best &  0.0014 &  3.8210 &  0.5502 &  {\bf 0.0008} &    0.0016 \\
     &         & mean &  {\bf 0.0023} &  4.1901 &  3.8734 &  0.4514 &    0.1321 \\
     &         & std &  {\bf 0.0014} &  0.2742 &  1.6595 &  1.3029 &    0.2132 \\
\bottomrule
\end{tabularx}
\label{table:compare_igd_all_30D}
\end{table*}

\begin{table*}[htbp]
\centering
\caption{Comparison of HV for the algorithms for $P=30$. The best results are highlighted. 
\label{Table:HV30}}
\begin{threeparttable}
\begin{tabularx}{0.65\linewidth}{l l l r r r r r}
\toprule
Eval \# & Instance & & MG-GPO & NSGA-II & MOPSO & MMOPSO & WOF-SMPSO \\
\midrule
1000 & ZDT1\_30 & best &  {\bf 0.5853} &  0.0068 &  0.0087 &  0.4429 &    0.0000 \\
     &         & mean &  {\bf 0.5507} &  0.0013 &  0.0009 &  0.2777 &    0.0000 \\
     &         & std &  0.0239 &  {\bf 0.0022} &  0.0027 &  0.1316 &    0.0000 \\
     & ZDT2\_30 & best &  {\bf 0.2811} &  0.0000 &  0.0000 &  0.0000 &    0.0000 \\
     &         & mean &  {\bf 0.2419} &  0.0000 &  0.0000 &  0.0000 &    0.0000 \\
     &         & std &  {\bf 0.0348} &  0.0000 &  0.0000 &  0.0000 &    0.0000 \\
     & ZDT3\_30 & best &  {\bf 0.7626} &  0.1955 &  0.0007 &  0.4505 &    0.0007 \\
     &         & mean &  {\bf 0.6371} &  0.1111 &  0.0001 &  0.2821 &    0.0001 \\
     &         & std &  0.1101 &  0.0483 &  0.0002 &  0.1033 &    {\bf 0.0002} \\
     & ZDT6\_30 & best &  0.0000 &  0.0000 &  0.0000 &  0.0000 &    0.0000 \\
     &         & mean &  0.0000 &  0.0000 &  0.0000 &  0.0000 &    0.0000 \\
     &         & std &  0.0000 &  0.0000 &  0.0000 &  0.0000 &    0.0000 \\
2000 & ZDT1\_30 & best &  {\bf 0.6608} &  0.2012 &  0.2468 &  0.6578 &    0.6047 \\
     &         & mean &  {\bf 0.6560} &  0.1528 &  0.0510 &  0.5637 &    0.4447 \\
     &         & std &  {\bf 0.0036} &  0.0302 &  0.0809 &  0.1240 &    0.0885 \\
     & ZDT2\_30 & best &  {\bf 0.3297} &  0.0000 &  0.0104 &  0.2539 &    0.3288 \\
     &         & mean &  {\bf 0.3284} &  0.0000 &  0.0010 &  0.0372 &    0.0666 \\
     &         & std &  {\bf 0.0011} &  0.0000 &  0.0033 &  0.0837 &    0.1120 \\
     & ZDT3\_30 & best &  {\bf 0.9768} &  0.6169 &  0.1645 &  0.9677 &    0.7900 \\
     &         & mean &  {\bf 0.9288} &  0.4877 &  0.0506 &  0.7488 &    0.5097 \\
     &         & std &  {\bf 0.0456} &  0.0712 &  0.0530 &  0.1236 &    0.2176 \\
     & ZDT6\_30 & best &  {\bf 0.1796} &  0.0000 &  0.0000 &  0.0000 &    0.0000 \\
     &         & mean &  {\bf 0.0410} &  0.0000 &  0.0000 &  0.0000 &    0.0000 \\
     &         & std &  {\bf 0.0693} &  0.0000 &  0.0000 &  0.0000 &    0.0000 \\
3000 & ZDT1\_30 & best &  0.6623 &  0.3862 &  0.4432 &  {\bf 0.6644} &    0.6471 \\
     &         & mean &  {\bf 0.6589} &  0.3118 &  0.1845 &  0.6407 &    0.5106 \\
     &         & std &  {\bf 0.0020} &  0.0432 &  0.1187 &  0.0601 &    0.0986 \\
     & ZDT2\_30 & best &  {\bf 0.3314} &  0.0562 &  0.1263 &  0.3300 &    0.3300 \\
     &         & mean &  {\bf 0.3311} &  0.0165 &  0.0426 &  0.0776 &    0.1439 \\
     &         & std &  {\bf 0.0003} &  0.0229 &  0.0524 &  0.1363 &    0.1327 \\
     & ZDT3\_30 & best &  1.0149 &  0.7924 &  0.2755 &  {\bf 1.0243} &    0.8527 \\
     &         & mean &  {\bf 0.9819} &  0.6759 &  0.1253 &  0.8847 &    0.6562 \\
     &         & std &  {\bf 0.0175} &  0.0735 &  0.0806 &  0.1025 &    0.1336 \\
     & ZDT6\_30 & best &  0.3220 &  0.0000 &  0.0000 &  {\bf 0.3237} &    0.1311 \\
     &         & mean &  {\bf 0.3112} &  0.0000 &  0.0000 &  0.0660 &    0.0219 \\
     &         & std &  {\bf 0.0168} &  0.0000 &  0.0000 &  0.1145 &    0.0457 \\
4000 & ZDT1\_30 & best &  0.6624 &  0.5089 &  0.5522 &  {\bf 0.6656} &    0.6587 \\
     &         & mean &  0.6597 &  0.4427 &  0.3301 &  {\bf 0.6630} &    0.5767 \\
     &         & std &  {\bf 0.0019} &  0.0435 &  0.1273 &  0.0033 &    0.1163 \\
     & ZDT2\_30 & best &  0.3320 &  0.1700 &  0.2479 &  0.3319 &    {\bf 0.3323} \\
     &         & mean &  {\bf 0.3318} &  0.0919 &  0.1257 &  0.1406 &    0.2411 \\
     &         & std &  {\bf 0.0002} &  0.0540 &  0.0962 &  0.1560 &    0.1368 \\
     & ZDT3\_30 & best &  1.0312 &  0.8851 &  0.4324 &  {\bf 1.0394} &    0.9594 \\
     &         & mean &  {\bf 1.0071} &  0.7877 &  0.3184 &  0.9663 &    0.8021 \\
     &         & std &  {\bf 0.0190} &  0.0703 &  0.0899 &  0.0890 &    0.1274 \\
     & ZDT6\_30 & best &  0.3244 &  0.0000 &  0.0472 &  {\bf 0.3251} &    0.3242 \\
     &         & mean &  {\bf 0.3232} &  0.0000 &  0.0047 &  0.2545 &    0.2384 \\
     &         & std &  {\bf 0.0019} &  0.0000 &  0.0149 &  0.1344 &    0.1318 \\
\bottomrule\addlinespace[1ex]
\end{tabularx}
\begin{tablenotes}\footnotesize
\item[$\dagger$] The reference point in the HV calculation for all the test instances is set to $(1, 1)$.
\item[*] The  zero HV value indicates that for all 10 runs {no solution has reached the reference point}.
\end{tablenotes}
\end{threeparttable}
\label{table:compare_hv_all_30D}
\end{table*}

\begin{table*}[htbp]
\centering
\caption{Comparison of IGD for $P=100$. The best results are highlighted.
\label{Table:IGD100}}
\begin{tabularx}{0.65\linewidth}{l l l r r r r r}
\toprule
Eval \# & Instance & & MG-GPO & NSGA-II & MOPSO & MMOPSO & WOF-SMPSO \\
\midrule
1000 & ZDT1\_100 & best &  {\bf 0.7408} &  1.5735 &  1.6825 &  0.9650 &    2.0264 \\
     &          & mean &  {\bf 0.7941} &  1.7136 &  1.8836 &  1.3174 &    2.1810 \\
     &          & std &  {\bf 0.0437} &  0.0945 &  0.1193 &  0.1793 &    0.1215 \\
     & ZDT2\_100 & best &  {\bf 0.9612} &  2.7032 &  3.0279 &  2.3379 &    2.8019 \\
     &          & mean &  {\bf 1.2484} &  2.8831 &  3.1692 &  3.0501 &    3.2418 \\
     &          & std &  0.1920 &  {\bf 0.1150} &  0.1267 &  0.4158 &    0.2422 \\
2000 & ZDT1\_100 & best &  {\bf 0.1908} &  0.8558 &  1.3619 &  0.5904 &    0.1959 \\
     &          & mean &  {\bf 0.2453} &  1.0487 &  1.5337 &  0.7818 &    0.3353 \\
     &          & std &  {\bf 0.0427} &  0.1117 &  0.1315 &  0.1047 &    0.0954 \\
     & ZDT2\_100 & best &  {\bf 0.1274} &  1.7740 &  2.4465 &  1.9072 &    0.6105 \\
     &          & mean &  {\bf 0.2524} &  1.9278 &  2.6609 &  2.7419 &    0.6105 \\
     &          & std &  0.1353 &  0.1350 &  0.1149 &  0.4854 &    {\bf 0.0000} \\
4000 & ZDT1\_100 & best &  {\bf 0.0131} &  0.5059 &  0.9513 &  0.2877 &    0.0244 \\
     &          & mean &  {\bf 0.0241} &  0.5997 &  1.1984 &  0.4597 &    0.0880 \\
     &          & std &  {\bf 0.0102} &  0.0758 &  0.1326 &  0.1299 &    0.0508 \\
     & ZDT2\_100 & best &  {\bf 0.0021} &  1.0871 &  1.8766 &  1.4243 &    0.4426 \\
     &          & mean &  {\bf 0.0046} &  1.2850 &  2.0804 &  2.4133 &    0.5801 \\
     &          & std &  {\bf 0.0032} &  0.1985 &  0.1329 &  0.5493 &    0.0645 \\
8000 & ZDT1\_100 & best &  {\bf 0.0020} &  0.2572 &  0.5027 &  0.1786 &    0.0030 \\
     &          & mean &  {\bf 0.0024} &  0.3542 &  0.8265 &  0.2476 &    0.0083 \\
     &          & std &  {\bf 0.0005} &  0.0681 &  0.1939 &  0.0636 &    0.0083 \\
     & ZDT2\_100 & best &  {\bf 0.0004} &  0.5963 &  1.2890 &  0.8107 &    0.0013 \\
     &          & mean &  {\bf 0.0006} &  0.7818 &  1.5314 &  1.7554 &    0.2945 \\
     &          & std &  {\bf 0.0001} &  0.1882 &  0.1977 &  0.5665 &    0.2641 \\
\bottomrule
\end{tabularx}
\label{table:compare_igd_all_100D}
\end{table*}

\begin{table*}[htbp]
\centering
\caption{Comparison of HV for the algorithms for $P=100$. The best results are highlighted. 
\label{Table:HV100}}
\begin{threeparttable}
\begin{tabularx}{0.65\linewidth}{l l l r r r r r}
\toprule
Eval \# & Instance & & MG-GPO & NSGA-II & MOPSO & MMOPSO & WOF-SMPSO \\
\midrule
1000 & ZDT1\_100 & best &  {\bf 0.0156} &  0.0000 &  0.0000 &  0.0000 &    0.0000 \\
     &          & mean &  {\bf 0.0054} &  0.0000 &  0.0000 &  0.0000 &    0.0000 \\
     &          & std &  {\bf 0.0055} &  0.0000 &  0.0000 &  0.0000 &    0.0000 \\
     & ZDT2\_100 & best &  0.0000 &  0.0000 &  0.0000 &  0.0000 &    0.0000 \\
     &          & mean &  0.0000 &  0.0000 &  0.0000 &  0.0000 &    0.0000 \\
     &          & std &  0.0000 &  0.0000 &  0.0000 &  0.0000 &    0.0000 \\
2000 & ZDT1\_100 & best &  0.3988 &  0.0008 &  0.0000 &  0.0526 &    {\bf 0.4345} \\
     &          & mean &  {\bf 0.3287} &  0.0001 &  0.0000 &  0.0107 &    0.3040 \\
     &          & std &  0.0449 &  {\bf 0.0003} &  0.0000 &  0.0171 &    0.0951 \\
     & ZDT2\_100 & best &  {\bf 0.1865} &  0.0000 &  0.0000 &  0.0000 &    0.0000 \\
     &          & mean &  {\bf 0.1103} &  0.0000 &  0.0000 &  0.0000 &    0.0000 \\
     &          & std &  {\bf 0.0539} &  0.0000 &  0.0000 &  0.0000 &    0.0000 \\
4000 & ZDT1\_100 & best &  {\bf 0.6452} &  0.1163 &  0.0000 &  0.2944 &    0.6365 \\
     &          & mean &  {\bf 0.6263} &  0.0661 &  0.0000 &  0.1558 &    0.5717 \\
     &          & std &  {\bf 0.0157} &  0.0343 &  0.0000 &  0.0900 &    0.0658 \\
     & ZDT2\_100 & best &  {\bf 0.3298} &  0.0000 &  0.0000 &  0.0000 &    0.0344 \\
     &          & mean &  {\bf 0.3256} &  0.0000 &  0.0000 &  0.0000 &    0.0058 \\
     &          & std &  {\bf 0.0050} &  0.0000 &  0.0000 &  0.0000 &    0.0124 \\
8000 & ZDT1\_100 & best &  {\bf 0.6624} &  0.3279 &  0.1173 &  0.4210 &    0.6621 \\
     &          & mean &  {\bf 0.6610} &  0.2381 &  0.0227 &  0.3418 &    0.6554 \\
     &          & std &  {\bf 0.0014} &  0.0606 &  0.0377 &  0.0713 &    0.0105 \\
     & ZDT2\_100 & best &  {\bf 0.3326} &  0.0007 &  0.0000 &  0.0000 &    0.3310 \\
     &          & mean &  {\bf 0.3322} &  0.0001 &  0.0000 &  0.0000 &    0.1499 \\
     &          & std &  {\bf 0.0002} &  0.0002 &  0.0000 &  0.0000 &    0.1496 \\
\bottomrule\addlinespace[1ex]
\end{tabularx}
\begin{tablenotes}\footnotesize
\item[$\dagger$] The reference point in the HV calculation for all the test instances is set to $(1, 1)$.
\item[*] The zero HV value has the same meaning as in Table~\ref{table:compare_hv_all_30D}.
\end{tablenotes}
\end{threeparttable}
\label{table:compare_hv_all_100D}
\end{table*}

\begin{table}[htbp]
\centering
\caption{Wilcoxon test results at a 0.05 significance level: IGD ($P=30$). The test is performed between MG-GPO and each of the other 4 algorithms.}
\begin{threeparttable}
\begin{tabularx}{\linewidth}{l l c c c c}
\toprule
Eval \# & Instance & NSGA-II & MOPSO & MMOPSO & WOF-SMPSO \\
\midrule
1000 & ZDT1\_30 &   1 &      1 &       1 &         1 \\
     & ZDT2\_30 &   1 &      1 &       1 &         1 \\
     & ZDT3\_30 &   1 &      1 &       1 &         1 \\
     & ZDT6\_30 &   1 &      1 &       1 &         1 \\
2000 & ZDT1\_30 &   1 &      1 &       1 &         1 \\
     & ZDT2\_30 &   1 &      1 &       1 &         1 \\
     & ZDT3\_30 &   1 &      1 &       1 &         1 \\
     & ZDT6\_30 &   1 &      1 &       1 &         0 \\
3000 & ZDT1\_30 &   1 &      1 &       0 &         1 \\
     & ZDT2\_30 &   1 &      1 &       1 &         1 \\
     & ZDT3\_30 &   1 &      1 &       0 &         1 \\
     & ZDT6\_30 &   1 &      1 &       1 &         1 \\
4000 & ZDT1\_30 &   1 &      1 &       0 &         1 \\
     & ZDT2\_30 &   1 &      1 &       1 &         1 \\
     & ZDT3\_30 &   1 &      1 &       0 &         1 \\
     & ZDT6\_30 &   1 &      1 &       0 &         1 \\
\bottomrule\addlinespace[1ex]
\end{tabularx}
\begin{tablenotes}\footnotesize
\item[*] 1 denotes that MG-GPO wins, 0 for no winner, -1 for that the other  algorithm wins.
\end{tablenotes}
\end{threeparttable}
\label{table:wilcoxon_igd_30D}
\end{table}

\begin{table}[htbp]
\centering
\caption{Wilcoxon test results at a 0.05 significance level: HV ($P=30$). The test is performed between MG-GPO and each of the other 4 algorithms.}
\begin{threeparttable}
\begin{tabularx}{\linewidth}{l l c c c c}
\toprule
Eval \# & Instance & NSGA-II & MOPSO & MMOPSO & WOF-SMPSO \\
\midrule
1000 & ZDT1\_30 &    1 &      1 &       1 &         1 \\
     & ZDT2\_30 &    1 &      1 &       1 &         1 \\
     & ZDT3\_30 &    1 &      1 &       1 &         1 \\
     & ZDT6\_30 &    N/A &    N/A &     N/A &       N/A \\
2000 & ZDT1\_30 &    1 &      1 &       0 &         1 \\
     & ZDT2\_30 &    1 &      1 &       1 &         1 \\
     & ZDT3\_30 &    1 &      1 &       1 &         1 \\
     & ZDT6\_30 &    0 &      0 &       0 &         0 \\
3000 & ZDT1\_30 &    1 &      1 &       0 &         1 \\
     & ZDT2\_30 &    1 &      1 &       1 &         1 \\
     & ZDT3\_30 &    1 &      1 &       0 &         1 \\
     & ZDT6\_30 &    1 &      1 &       1 &         1 \\
4000 & ZDT1\_30 &    1 &      1 &       -1 &        1 \\
     & ZDT2\_30 &    1 &      1 &       1 &         1 \\
     & ZDT3\_30 &    1 &      1 &       0 &         1 \\
     & ZDT6\_30 &    1 &      1 &       0 &         0 \\
\bottomrule\addlinespace[1ex]
\end{tabularx}
\begin{tablenotes}\footnotesize
\item[*] 1, 0, and -1 have the same meaning as in Table~\ref{table:wilcoxon_igd_30D}. 
N/A denotes that neither algorithm has reached a positive HV among the 10 runs.
\end{tablenotes}
\end{threeparttable}
\label{table:wilcoxon_hv_30D}
\end{table}

\begin{table}[htbp]
\centering
\caption{Wilcoxon test results at a 0.05 significance level: IGD ($P=100$). The test is performed between MG-GPO and each of the other 4 algorithms. 
}
\begin{tabularx}{\linewidth}{l l c c c c}
\toprule
Eval \# & Instance & NSGA-II & MOPSO & MMOPSO & WOF-SMPSO \\
\midrule
1000 & ZDT1\_100 &   1 &      1 &       1 &         1 \\
     & ZDT2\_100 &   1 &      1 &       1 &         1 \\
2000 & ZDT1\_100 &   1 &      1 &       1 &         0 \\
     & ZDT2\_100 &   1 &      1 &       1 &         1 \\
4000 & ZDT1\_100 &   1 &      1 &       1 &         1 \\
     & ZDT2\_100 &   1 &      1 &       1 &         1 \\
8000 & ZDT1\_100 &   1 &      1 &       1 &         1 \\
     & ZDT2\_100 &   1 &      1 &       1 &         1 \\
\bottomrule\addlinespace[1ex]
\end{tabularx}
\begin{tablenotes}\footnotesize
\item[*] 1, 0, and -1 have the same meaning as in Table~\ref{table:wilcoxon_igd_30D}. 
\end{tablenotes}
\label{table:wilcoxon_igd_100D}
\end{table}

\begin{table}[htbp]
\centering
\caption{Wilcoxon test results at a 0.05 significance level: HV ($P=100$). The test is performed between MG-GPO and each of the other 4 algorithms.}
\begin{threeparttable}
\begin{tabularx}{\linewidth}{l l c c c c}
\toprule
Eval \# & Instance & NSGA-II & MOPSO & MMOPSO & WOF-SMPSO \\
\midrule
1000 & ZDT1\_100 &    1 &      1 &       1 &         1 \\
     & ZDT2\_100 &    N/A &    N/A &     N/A &       N/A \\
2000 & ZDT1\_100 &    1 &      1 &       1 &         0 \\
     & ZDT2\_100 &    1 &      1 &       1 &         1 \\
4000 & ZDT1\_100 &    1 &      1 &       1 &         1 \\
     & ZDT2\_100 &    1 &      1 &       1 &         1 \\
8000 & ZDT1\_100 &    1 &      1 &       1 &         1 \\
     & ZDT2\_100 &    1 &      1 &       1 &         1 \\
\bottomrule\addlinespace[1ex]
\end{tabularx}
\begin{tablenotes}\footnotesize
\item[*] 1, 0, -1 and N/A have the same meaning as in Table~\ref{table:wilcoxon_hv_30D}.
\end{tablenotes}
\end{threeparttable}
\label{table:wilcoxon_hv_100D}
\end{table}



\section{Conclusion \label{secConcl}}
We proposed a new multi-objective stochastic optimization algorithm 
that is based on Gaussian process regression. 
The new algorithm update a population of solutions iteratively. 
At each iteration, it constructs a posterior Gaussian process and uses 
it as a surrogate model of the actual system to be optimized. 
A large number of candidate solutions are generated and 
evaluated with the surrogate model and the results are used to select 
a small number of promising solutions to be evaluated on the real system (e.g., 
by physics simulation). 

The new algorithm, referred to as multi-generation Gaussian process optimizer (MG-GPO), 
has been tested with analytic functions. 
In all test cases, a substantially faster convergence speed is found than the classic 
evolutionary algrithms, 
NSGA-II and MOPSO, as well as two more recent algorithms, MMOPOSO and WOF-SMPSO. 
The new algorithm would be very useful for design optimization of large systems
where a search for optimal solutions in a multi-dimensional parameter space is 
needed. 

Future improvement could be made to the MG-GPO algorithm by adopting an adaptive 
scheme to update the $\kappa$ hyper-parameter for the GP-LCB acquisition function. 

\section*{Acknowledgment}
  This work was supported by the U.S. Department of Energy, Office of
  Science, Office of Basic Energy Sciences, under Contract No.
  DE-AC02-76SF00515 and FWP 2018-SLAC-100469 and 
  by Computing Science, Office of Advanced Scientific Computing Research under FWP 
  2018-SLAC-100469ASCR.

\ifCLASSOPTIONcaptionsoff
  \newpage
\fi



%
\bibliographystyle{IEEEtran}
\bibliography{mggpo.bib}

\begin{thebibliography}{10}
\providecommand{\url}[1]{#1}
\csname url@samestyle\endcsname
\providecommand{\newblock}{\relax}
\providecommand{\bibinfo}[2]{#2}
\providecommand{\BIBentrySTDinterwordspacing}{\spaceskip=0pt\relax}
\providecommand{\BIBentryALTinterwordstretchfactor}{4}
\providecommand{\BIBentryALTinterwordspacing}{\spaceskip=\fontdimen2\font plus
\BIBentryALTinterwordstretchfactor\fontdimen3\font minus
  \fontdimen4\font\relax}
\providecommand{\BIBforeignlanguage}[2]{{%
\expandafter\ifx\csname l@#1\endcsname\relax
\typeout{** WARNING: IEEEtran.bst: No hyphenation pattern has been}%
\typeout{** loaded for the language `#1'. Using the pattern for}%
\typeout{** the default language instead.}%
\else
\language=\csname l@#1\endcsname
\fi
#2}}
\providecommand{\BIBdecl}{\relax}
\BIBdecl

\bibitem{DebGAbook}
K.~Deb and D.~Kalyanmoy, \emph{Multi-Objective Optimization Using Evolutionary
  Algorithms}.\hskip 1em plus 0.5em minus 0.4em\relax New York, NY, USA: John
  Wiley \& Sons, Inc., 2001.

\bibitem{NSGA2}
K.~{Deb}, A.~{Pratap}, S.~{Agarwal}, and T.~{Meyarivan}, ``A fast and elitist
  multiobjective genetic algorithm: Nsga-ii,'' \emph{IEEE Transactions on
  Evolutionary Computation}, vol.~6, no.~2, pp. 182--197, April 2002.

\bibitem{MOEAD}
Q.~{Zhang} and H.~{Li}, ``Moea/d: A multiobjective evolutionary algorithm based
  on decomposition,'' \emph{IEEE Transactions on Evolutionary Computation},
  vol.~11, no.~6, pp. 712--731, 2007.

\bibitem{Kennedy488968}
J.~{Kennedy} and R.~{Eberhart}, ``Particle swarm optimization,'' in
  \emph{Proceedings of ICNN'95 - International Conference on Neural Networks},
  vol.~4, Nov 1995, pp. 1942--1948 vol.4.

\bibitem{PSO985692Kennedy}
M.~{Clerc} and J.~{Kennedy}, ``The particle swarm - explosion, stability, and
  convergence in a multidimensional complex space,'' \emph{IEEE Transactions on
  Evolutionary Computation}, vol.~6, no.~1, pp. 58--73, Feb 2002.

\bibitem{MMOPSO}
Q.~Lin, J.~Li, Z.~Du, J.~Chen, and Z.~Ming, ``A novel multi-objective particle
  swarm optimization with multiple search strategies,'' \emph{European Journal
  of Operational Research}, vol. 247, no.~3, pp. 732--744, 2015.

\bibitem{WOFSMPSO2018}
H.~{Zille}, H.~{Ishibuchi}, S.~{Mostaghim}, and Y.~{Nojima}, ``A framework for
  large-scale multiobjective optimization based on problem transformation,''
  \emph{IEEE Transactions on Evolutionary Computation}, vol.~22, no.~2, pp.
  260--275, 2018.

\bibitem{YangEPAC08MOGA}
L.~Yang, D.~Robin, F.~Sannibale, C.~Steier, and W.~Wan, ``Global optimization
  of the magnetic lattice using genetic algorithms,'' in \emph{Proceedings of
  EPAC'08}, 2008, pp. 3050--3052.

\bibitem{BorlandPAC09DA}
M.~Borland, V.~Sajaev, L.~Emery, and A.~Xiao, ``Direct methods of optimization
  of storage ring dynamic and momentum aperture,'' in \emph{Proceedings of
  PAC'09}, 2009, pp. 3850--3852.

\bibitem{HUANG201448PSO}
\BIBentryALTinterwordspacing
X.~Huang and J.~Safranek, ``Nonlinear dynamics optimization with particle swarm
  and genetic algorithms for spear3 emittance upgrade,'' \emph{Nuclear
  Instruments and Methods in Physics Research Section A: Accelerators,
  Spectrometers, Detectors and Associated Equipment}, vol. 757, pp. 48 -- 53,
  2014. [Online]. Available:
  \url{http://www.sciencedirect.com/science/article/pii/S0168900214004914}
\BIBentrySTDinterwordspacing

\bibitem{BazarovMOGA}
\BIBentryALTinterwordspacing
I.~V. Bazarov and C.~K. Sinclair, ``Multivariate optimization of a high
  brightness dc gun photoinjector,'' \emph{Phys. Rev. ST Accel. Beams}, vol.~8,
  p. 034202, Mar 2005. [Online]. Available:
  \url{https://link.aps.org/doi/10.1103/PhysRevSTAB.8.034202}
\BIBentrySTDinterwordspacing

\bibitem{PANG2014124PSO}
\BIBentryALTinterwordspacing
X.~Pang and L.~Rybarcyk, ``Multi-objective particle swarm and genetic algorithm
  for the optimization of the lansce linac operation,'' \emph{Nuclear
  Instruments and Methods in Physics Research Section A: Accelerators,
  Spectrometers, Detectors and Associated Equipment}, vol. 741, pp. 124 -- 129,
  2014. [Online]. Available:
  \url{http://www.sciencedirect.com/science/article/pii/S0168900213017464}
\BIBentrySTDinterwordspacing

\bibitem{MOEADEGO}
Q.~{Zhang}, W.~{Liu}, E.~{Tsang}, and B.~{Virginas}, ``Expensive multiobjective
  optimization by moea/d with gaussian process model,'' \emph{IEEE Transactions
  on Evolutionary Computation}, vol.~14, no.~3, pp. 456--474, 2010.

\bibitem{KRVEA}
T.~{Chugh}, Y.~{Jin}, K.~{Miettinen}, J.~{Hakanen}, and K.~{Sindhya}, ``A
  surrogate-assisted reference vector guided evolutionary algorithm for
  computationally expensive many-objective optimization,'' \emph{IEEE
  Transactions on Evolutionary Computation}, vol.~22, no.~1, pp. 129--142,
  2018.

\bibitem{ParEGO}
J.~{Knowles}, ``{ParEGO}: a hybrid algorithm with on-line landscape
  approximation for expensive multiobjective optimization problems,''
  \emph{IEEE Transactions on Evolutionary Computation}, vol.~10, no.~1, pp.
  50--66, 2006.

\bibitem{Kushner_1964GP}
\BIBentryALTinterwordspacing
H.~J. Kushner, ``A new method of locating the maximum point of an arbitrary
  multipeak curve in the presence of noise,'' \emph{Journal of Basic
  Engineering}, vol.~86, no.~1, p.~97, 1964. [Online]. Available:
  \url{https://doi.org/10.11152F1.3653121}
\BIBentrySTDinterwordspacing

\bibitem{ZhilinskasGP}
A.~G.~Zhilinskas, ``Single-step bayesian search method for an extremum of
  functions of a single variable,'' \emph{Cybernetics and Systems Analysis -
  CYBERN SYST ANAL-ENGL TR}, vol.~11, pp. 160--166, 01 1975.

\bibitem{Mockus77GP}
\BIBentryALTinterwordspacing
J.~Mockus, ``On bayesian methods for seeking the extremum and their
  application.'' in \emph{IFIP Congress}, 1977, pp. 195--200. [Online].
  Available:
  \url{http://dblp.uni-trier.de/db/conf/ifip/ifip1977.html\#Mockus77}
\BIBentrySTDinterwordspacing

\bibitem{Jones1998GP}
\BIBentryALTinterwordspacing
D.~R. Jones, M.~Schonlau, and W.~J. Welch, ``Efficient global optimization of
  expensive black-box functions,'' \emph{Journal of Global Optimization},
  vol.~13, no.~4, pp. 455--492, Dec 1998. [Online]. Available:
  \url{https://doi.org/10.1023/A:1008306431147}
\BIBentrySTDinterwordspacing

\bibitem{RasmussenGP}
C.~E. Rasmussen and C.~K.~I. Williams, \emph{Gaussian Processes for Machine
  Learning}.\hskip 1em plus 0.5em minus 0.4em\relax the MIT Press, 2006.

\bibitem{BrochuGPTut2010}
\BIBentryALTinterwordspacing
E.~Brochu, V.~M. Cora, and N.~de~Freitas, ``A tutorial on bayesian optimization
  of expensive cost functions, with application to active user modeling and
  hierarchical reinforcement learning,'' \emph{CoRR}, vol. abs/1012.2599, 2010.
  [Online]. Available: \url{http://arxiv.org/abs/1012.2599}
\BIBentrySTDinterwordspacing

\bibitem{Zitzler2000}
\BIBentryALTinterwordspacing
E.~Zitzler, K.~Deb, and L.~Thiele, ``Comparison of multiobjective evolutionary
  algorithms: Empirical results,'' \emph{Evol. Comput.}, vol.~8, no.~2, pp.
  173--195, Jun. 2000. [Online]. Available:
  \url{http://dx.doi.org/10.1162/106365600568202}
\BIBentrySTDinterwordspacing

\bibitem{curseofdim}
\BIBentryALTinterwordspacing
R.~Bellman, R.~Corporation, and K.~M.~R. Collection, \emph{Dynamic
  Programming}, ser. Rand Corporation research study.\hskip 1em plus 0.5em
  minus 0.4em\relax Princeton University Press, 1957. [Online]. Available:
  \url{https://books.google.it/books?id=wdtoPwAACAAJ}
\BIBentrySTDinterwordspacing

\bibitem{Syberfeldt2008}
A.~{Syberfeldt}, H.~{Grimm}, A.~{Ng}, and R.~I. {John}, ``A parallel
  surrogate-assisted multi-objective evolutionary algorithm for computationally
  expensive optimization problems,'' in \emph{2008 IEEE Congress on
  Evolutionary Computation (IEEE World Congress on Computational
  Intelligence)}, 2008, pp. 3177--3184.

\bibitem{Liu2008}
P.-G. Liu, X.~Han, and C.~Jiang, ``A novel multi-objective optimization method
  based on an approximation model management technique,'' \emph{Computer
  Methods in Applied Mechanics and Engineering}, vol. 197, pp. 2719--2731, 06
  2008.

\bibitem{Kourakos2013}
G.~Kourakos and A.~Mantoglou, ``Development of a multi-objective optimization
  algorithm using surrogate models for coastal aquifer management,''
  \emph{Journal of Hydrology}, vol. 479, p. 13–23, 02 2013.

\bibitem{EdelenML}
A.~Edelen, A.~Adelmann, N.~Neveu, Y.~Huber, and M.~Frey, ``Machine learning to
  enable orders of magnitude speedup in multi-objective optimization of
  particle accelerator systems,'' \url{https://arxiv.org/pdf/1903.07759.pdf},
  March 2019.

\bibitem{GPUCB}
P.~Auer, ``Using confidence bounds forexploitation-exploration trade-offs,''
  \emph{ournal of Machine Learning Research}, vol.~3, pp. 397--422, 2002.

\bibitem{DebSBX1995}
K.~Deb and R.~B. Agrawal, ``Simulated binary crossover for continuous search
  space,'' \emph{Complex Systems}, vol.~9, pp. 115--148, 1995.

\bibitem{liagkouras2013elitist}
K.~Liagkouras and K.~Metaxiotis, ``An elitist polynomial mutation operator for
  improved performance of moeas in computer networks,'' in \emph{2013 22nd
  International Conference on Computer Communication and Networks
  (ICCCN)}.\hskip 1em plus 0.5em minus 0.4em\relax IEEE, 2013, pp. 1--5.

\bibitem{WOFSMPSO2016}
\BIBentryALTinterwordspacing
H.~Zille, H.~Ishibuchi, S.~Mostaghim, and Y.~Nojima, ``Weighted optimization
  framework for large-scale multi-objective optimization,'' in
  \emph{Proceedings of the 2016 on Genetic and Evolutionary Computation
  Conference Companion}, ser. GECCO ’16 Companion.\hskip 1em plus 0.5em minus
  0.4em\relax New York, NY, USA: Association for Computing Machinery, 2016, p.
  83–84. [Online]. Available: \url{https://doi.org/10.1145/2908961.2908979}
\BIBentrySTDinterwordspacing

\bibitem{WOFSMPSO2017}
H.~{Zille} and S.~{Mostaghim}, ``Comparison study of large-scale optimisation
  techniques on the lsmop benchmark functions,'' in \emph{2017 IEEE Symposium
  Series on Computational Intelligence (SSCI)}, 2017, pp. 1--8.

\bibitem{PlatEMO}
Y.~Tian, R.~Cheng, X.~Zhang, and Y.~Jin, ``{PlatEMO}: A {MATLAB} platform for
  evolutionary multi-objective optimization,'' \emph{IEEE Computational
  Intelligence Magazine}, vol.~12, no.~4, pp. 73--87, 2017.

\bibitem{gpy2014}
{GPy}, ``{GPy}: A gaussian process framework in python,''
  \url{http://github.com/SheffieldML/GPy}, since 2012.

\bibitem{zitzler2003performance}
E.~Zitzler, L.~Thiele, M.~Laumanns, C.~M. Fonseca, and V.~G. Da~Fonseca,
  ``Performance assessment of multiobjective optimizers: An analysis and
  review,'' \emph{IEEE Transactions on evolutionary computation}, vol.~7,
  no.~2, pp. 117--132, 2003.

\bibitem{coello2004study}
C.~A.~C. Coello and M.~R. Sierra, ``A study of the parallelization of a
  coevolutionary multi-objective evolutionary algorithm,'' in \emph{Mexican
  International Conference on Artificial Intelligence}.\hskip 1em plus 0.5em
  minus 0.4em\relax Springer, 2004, pp. 688--697.

\bibitem{sierra2004new}
M.~R. Sierra and C.~A.~C. Coello, ``A new multi-objective particle swarm
  optimizer with improved selection and diversity mechanisms,'' \emph{Technical
  Report of CINVESTAV-IPN}, 2004.

\end{thebibliography}




%


\begin{IEEEbiographynophoto}{Xiaobiao Huang}
Xiaobiao Huang graduated from Tsinghua University with a BS in physics and a BE in 
computer science. He obtained a PhD degree in Physics from Indiana University, Bloomington 
in 2005. He has been a staff scientist at SLAC National Accelerator Laboratory since 
2006. 
\end{IEEEbiographynophoto}

\begin{IEEEbiographynophoto}{Minghao Song}
Minghao Song graduated from Chongqing University with a BE in 
Nuclear Engineering and Technology. He obtained a ME degree in Nuclear Energy and Nuclear Technology Engineering from University of Chinese Academy of Sciences. He is currently a Ph. D student in physics from Illinois Institute of Technology, Chicago and conducting Ph. D thesis work at SLAC National Accelerator Laboratory. 
\end{IEEEbiographynophoto}

\begin{IEEEbiographynophoto}{Zhe Zhang}
Zhe Zhang graduated from Tsinghua University with a BE in engineering physics. He obtained a PhD degree in nuclear science and technology from Tsinghua University in 2017. He is currently a research associate at SLAC National Accelerator Laboratory.
\end{IEEEbiographynophoto}




\end{document}